\documentclass[a4paper,twoside]{article}

\usepackage{tikz}

\usepackage{url}
\usepackage{float}
\usetikzlibrary{shapes,arrows,positioning}
\usepackage[plain]{algorithm}
\usepackage{mathtools,algorithm}
\usepackage[noend]{algpseudocode}
\usepackage{algpascal}
\usepackage{verbatim}
\usepackage{epsfig}
\usepackage{subcaption}
\usepackage{calc}
\usepackage{amssymb}
\usepackage{amstext}
\usepackage{amsmath}
\usepackage{amsthm}
\usepackage{multicol}
\usepackage{pslatex}
\usepackage{apalike}
\usepackage{color}
\usepackage{soul}
\usepackage{SCITEPRESS}     

\begin{document}
\title{\vspace*{-0.29cm}Exploiting Physical Contacts for Robustness Improvement of a Dot-Painting Mission by a Micro Air Vehicle}

\author{\authorname{Thomas Chaffre\sup{1}, Kevin Tudal\sup{1}, Sylvain Bertrand\sup{2} and Lionel Prevost\sup{1}}
\affiliation{\sup{1}Learning, Data and Robotics Lab, ESIEA, Paris, France}
\email{lionel.prevost@esiea.fr; thomas.chaffre@et.esiea.fr; kevin.tudal@et.esiea.fr} 
\affiliation{\sup{2}ONERA - The French Aerospace Lab, Palaiseau, France}
\email{sylvain.bertrand@onera.fr}}


\abstract{In this paper we address the problem of dot painting on a wall by a quadrotor Micro Air Vehicle (MAV), using on-board low cost sensors (monocular camera and IMU) for localization. A method is proposed to cope with uncertainties on the initial positioning of the MAV with respect to the wall and to deal with walls composed of multiple segments. This method is based on an online estimation algorithm that makes use of information of physical contacts detected by the drone during the flight to improve the positioning accuracy of the painted dots. Simulation results are presented to assess quantitatively the efficiency of the proposed approaches.}

\onecolumn \maketitle \normalsize \setcounter{footnote}{0} \vfill

\section{\uppercase{Introduction}}
\label{sec:introduction}

\noindent Robotics has experienced an outstanding growth and gained focus in recent years both from research and industry. Nowadays, use of robots in everyday life is becoming more and more usual. Researchers, teachers, students and artists are trying to use robotic platforms for creation, expression and sharing, bridging the gap between arts and engineering. As a matter of fact, Science, Technology, Engineering, Art and Mathematics (STEAM) are now considered as a whole in education or research, and in close relationship to robotics.\\
\indent The first cybernetic sculpture of history, named CYSP1, was created in 1956 by Nicolas Sch\"offer \cite{Pagliarini2009}. An electronic ``brain" attached to it allowed its sensors (photocells and microphone) to catch all natural variations in colour, light and noise intensity in its surroundings or made by the audience and therefore react to it. Ever since this period, a lot of attempts have been made to use robotic systems in the process of artistic creation, either autonomously or teleoperated by artists. Designing and disposing of a robot with drawing or painting capability is a challenge from a technical point of view, but it is also very interesting for artists in terms of creativity.
\\
\indent If the robotic system is supposed to duplicate a given drawing, painting is a process that requires multiple abilities: being able to perform precise movements, stay in contact with a surface and adapt to the environment changes. Aerial robots with Vertical Take-Off and Landing (VTOL) capabilities are appealing platforms thanks to the size of the operating volume enabled by aerial capacity and their low-speed and stationary flight capabilities. The term ``Painting Drone" appeared in the early 2014s \cite{Handy}. From this time forth, several painting drone projects where a quadcopter is equipped with a remotely activated spray-paint have been developed. One of the first to do that was the artist KATSU in 2015 who used a quadcopter to draw on a public billboard in the city of New-York. The firsts tangible projects of drone painting then emerged in 2016 with \cite{Pantograph} where an AR Drone 2.0 quadrotor was used to reproduce in real time the movements made by an user with a pen, and with \cite{Dot1} \cite{Dot2} where a small quadrotor robot was utilized for stippling. A motion capture system was used to measure the position of the robot and the canvas, and a robust control algorithm to drive the robot to different stipple positions and make contact with the canvas using an ink soaked sponge. In 2017, the exhibition ``Evolution 2.1" by Misha Most at the Winzavod Cultural Center in Moscow presented a fresco realized by a painting drone \cite{Misha}.
The quadcopter robot was able to precisely and autonomously paint on wide facades indoor and outdoor which can be considered as the best performance with a painting drone known to date.
\\
\indent In all the aforementioned works, the painting drones were either remotely controlled by the artist, or made use of an external motion capture system to localize themselves and perform the painting autonomously. This kind of setup can be expensive and restricts the use of these robots to known and adequately equipped environments. The problem investigated in this paper is therefore the one of autonomous dot-painting by a quadrotor using on-board sensors for localization.
\\
\indent In the process of dot-painting, contacts have to be made with the support to be painted (eg. wall, facade). Contrary to classical mobile robotics, where collision avoidance with the environment is a major concern, contact is here imposed by the mission. Use of collisions for the navigation of a flying robot has been addressed by some works of the literature. 
In \cite{Briod} a contact-based navigation method has been developed using a flying robot that can sustain collisions and that can self-recover once on the ground thanks to active legs. Random exploration is performed by the robot which taking advantage of the information from contact sensors to correct its direction after every collision with obstacles (walls, ceiling, floor). 
More recently, \cite{Franchi1} and \cite{Franchi2} have proposed control approaches to ensure physical interactions between a multi-rotor micro aerial vehicle and its environment for industrial use cases.\\
\indent In this paper, an estimation and control strategy is proposed to enable autonomous and accurate dot-painting missions by a quadrotor robot using only on-board low-cost sensors (monocular camera and IMU). 
The main contribution consists in the design of an online estimation procedure taking advantage of collisions to ensure robustness of the mission with respect to two types of uncertainties: uncertainty on the initial positioning of the robot with respect to the support that has to be painted; uncertainty on the knowledge of the shape of the support. Implementation and performance analysis of the proposed approach are realized through realistic simulations (ROS+Gazebo) by considering scenarios with both types of uncertainties.
\\
\indent The paper is organized as follows. Section 2 is devoted to notations and problem definition. Section 3 presents the navigation system and the dot-painting strategy. Section 4 presents the proposed estimation and control approach exploiting contact information and illustrate its efficiency on the scenario of dot-painting on a flat wall. Section 5 presents its extension to deal with walls composed of multiple segments. Finally, Section 6 concludes this work.

\section{Notations and problem  definition}

\subsection{Reference frames}

\begin{figure}[!h]
  \centering
   {\epsfig{file = 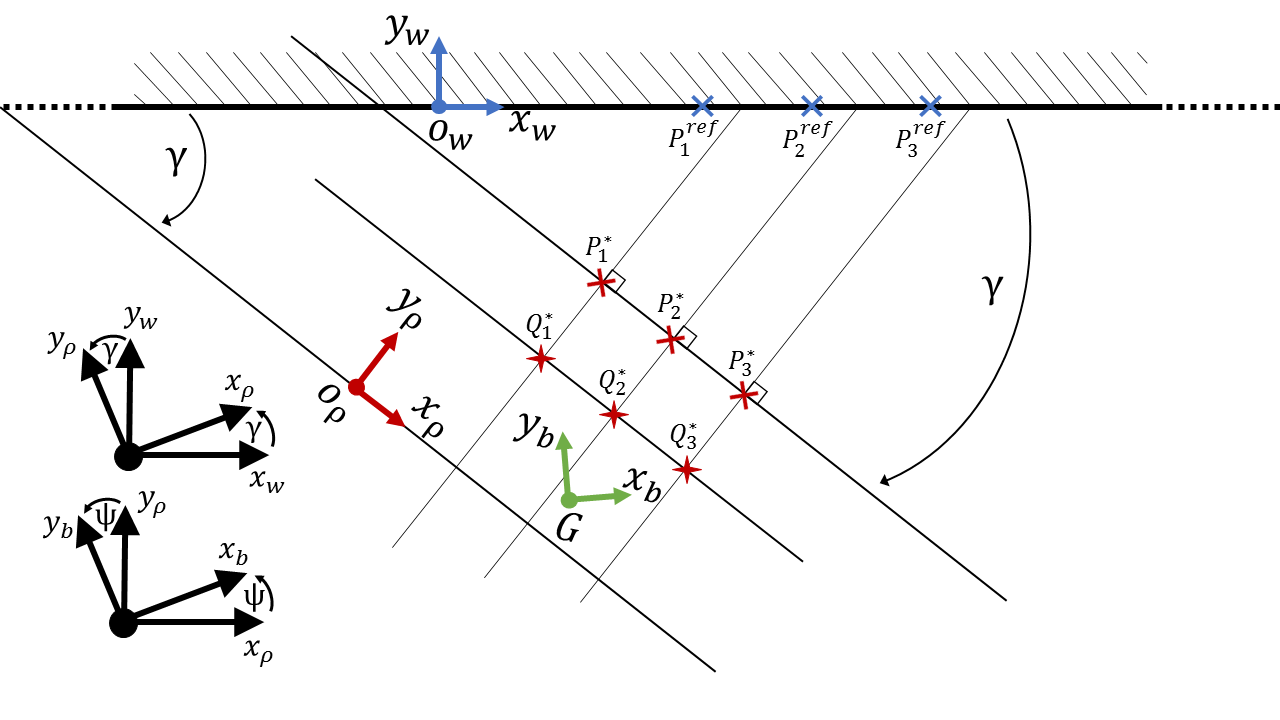, width = 1\columnwidth}}
  \caption{Dot-painting problem formalization.}
  \label{fig:notations}
  \vspace{-0.1cm}
\end{figure}

Three reference frames that will be used in the problem description are introduced in this section. They are presented in Figure \ref{fig:notations}. 
\\
\indent The first one is a fixed reference frame \( \mathcal{R}_w: (O_w; x_w, y_w, z_w)\) attached to the wall. Paint dots to be applied will be defined in this frame. 
\\
\indent The second frame is defined by \( \mathcal{R}_\rho: (O_\rho; x_\rho, y_\rho, z_\rho)\) and corresponds to the frame in which the localization of the robot is performed by its on-board navigation system. Its origin $O_\rho$ corresponds to the initial position of the quadrotor, before take-off, and its orientation is defined by the initial directions of the main axis of the quadrotor, before take-off. Assuming a 2D representation of the problem, the transformation $\mathcal{R}_w \rightarrow \mathcal{R}_\rho$ is defined by the translation vector $\delta = \overrightarrow{O_wO_\rho}$ and the rotation matrix $R(\gamma)$ parameterized by the angle $\gamma$. 
\\
\indent The third frame to be defined is the body frame \(\mathcal{R}_b: (G; x_b, y_b, z_b)\) attached to the center of gravity $G$ of the quadrotor robot and oriented along its main axis. The current position vector of the drone is denoted by $\xi = \overrightarrow{O_\rho G}$. The on-board navigation system of the drone will provide its position coordinates $\xi^{(\rho)}$ in $\mathcal{R}_\rho$. Assuming quasi-stationary flight (low speed and small attitude angles), and for simplification of the dot-painting problem, only the yaw angle $\psi$ of the quadrotor will be considered. The transformation $\mathcal{R}_\rho \rightarrow \mathcal{R}_b$ is therefore determined by the translation vector $\xi$ and the rotation matrix $R(\psi)$ parametrized by the angle $\psi$. Note that, before take-off the two frames $\mathcal{R}_\rho$ and $\mathcal{R}_b$ coincide. 
\subsection{Definition of paint dots}

The objective is to apply a set of $n \in \mathbb{N}^*$ paint dots \(\left\{ P_i^{ref(w)} \right\}\), $i=1,..,n$, whose coordinates are defined in $\mathcal{R}_w$ by 
\begin{gather*}
{P_i^{ref(w)}={
                \begin{bmatrix}
                    x_i^{ref(w)} \\
                    y_i^{ref(w)} \\
                    z_i^{ref(w)}
                \end{bmatrix}_w},\hspace{2.5mm} i = 1,\dots,n}
\end{gather*}
At the initial instant $t=0$, it is assumed that the drone disposes of a list of these points. Since localization information on the drone is available in $\mathcal{R}_\rho$, the coordinates of these points must therefore be defined in this frame. They are denoted by \( P_i^{*(\rho)}\) and defined by 
\begin{equation} \label{eq:def_pts_parfaite}
	P_i^{*(\rho)} = \begin{bmatrix}
                    x_i^{*(\rho)} \\
                    y_i^{*(\rho)} \\
                    z_i^{*(\rho)}
                \end{bmatrix}_\rho = R(\gamma)^T \left( P_i^{ref(w)} - \delta^{(w)} \right), \, i=1, .., n
\end{equation}
Note that this definition requires a perfect knowledge of $(\delta, \gamma)$. Uncertainties on these parameters will lead to errors in the resulting positions of the paint dots.
Although the whole dot-painting mission will be realized autonomously by the quadrotor, the initial positioning of the robot with respect to the wall is assumed to be done by a human operator. 
Knowing the desired location on the wall of the paint dots, the operator will be asked to place the drone facing the wall and at a given distance $d$ from the origin $O_w$\footnote{$O_w$ can be defined for example as the ground projection of the first point to be painted.}. If the accuracy of this initial positioning was perfect, one would have at $t=0$: $\delta^{(w)} = \xi^{(w)}(0) = [0, -d, 0]^T$ and $\gamma=\psi(0)=0$. Since in practice this initial positioning will suffer from uncertainties, an estimate of the value of $(\delta, \gamma)$ has to be performed and used to update the definition of the points $P_i^{*(\rho)}$ so that the pattern obtained by the paint dots matches the desired one as much as possible. 
\\
\indent This paper aims at proposing such a method to compute an online estimate $(\hat{\delta}^{(w)}, \hat{\gamma})$ of $(\delta^{(w)}, \gamma)$, along with the corresponding updates of the $P_i^{*(\rho)}$,  by exploiting information of physical contacts of the MAV with the wall. This information consists in the coordinates of the contact points denoted
\begin{gather*}
{P_i^{c(\rho)}={
                \begin{bmatrix}
                    x_i^{c(\rho)} \\
                    y_i^{c(\rho)} \\
                    z_i^{c(\rho)}
                \end{bmatrix}_\rho},\hspace{2.5mm} i = 1,\dots,n }
\end{gather*}
which correspond to the coordinates of the MAV provided in $\mathcal{R_\rho}$ by the on-board localization system when a contact with the wall is detected (see Section \ref{impact}). 
\\
\indent At the initial instant $t=0$, the initial value of the estimates are defined as $(\hat{\delta}^{(w)}(0), \hat{\gamma}(0))=([0, -d, 0]^T, 0)$ which corresponds to the instruction given to the operator for the quadrotor initial positioning. Instead of using \eqref{eq:def_pts_parfaite} which requires a perfect knowledge of $(\delta^{(w)}, \gamma)$ and which is not available, the list of paint dots given to the quadrotor at $t=0$ is computed using
\begin{equation}
	P_i^{*(\rho)} = R(\hat{\gamma}(0))^T \left( P_i^{ref(w)} - \hat{\delta}^{(w)}(0) \right), \, i=1, .., n
\end{equation}
As will be described later in the paper, accumulating information as contacts occur will improve the estimation process and the accuracy of the positioning of the next points to be painted. Detection of the contacts is explained in the next section along with localization and control strategy for the drone.

\section{Control and estimation architecture for dot-painting}

\noindent This section describes the localization and control architecture of the drone as well as the simulation tools used for implementation and validation of the proposed approach.

\subsection{Simulation environment} 

\noindent The proposed methods have been implemented using the ROS (Robot Operating System) middleware and the Gazebo simulator. The drone simulated is an AR Drone 2.0 and the \textit{ardrone\_autonomy} \cite{AR} and \textit{tum\_ardrone} \cite{Tum} packages are used. Gazebo is a 3D dynamic simulator which offers physics simulation with a high degree of fidelity and a large suite of robot and sensor models. The model used to simulate the quadrotor robot is representative of the true robot dynamics but does not account for external perturbations. In reality, the air flow generated by the rotors acts as an external perturbation that can disturb the quadcopter flight when it is operating close to a facade. This is not included in our simulations and will be addressed in future work.\\
\indent To mimic the act of dot-painting in Gazebo, we decided to make the quadcopter touch the facade and then to make a dot appears (represented by the yellow dots in Figure 3) at the collision coordinates. 

\subsection{Contact painting strategy} \label{impact}
\begin{figure}[!h]
  \vspace{-0.2cm}
  \centering
   {\epsfig{file = 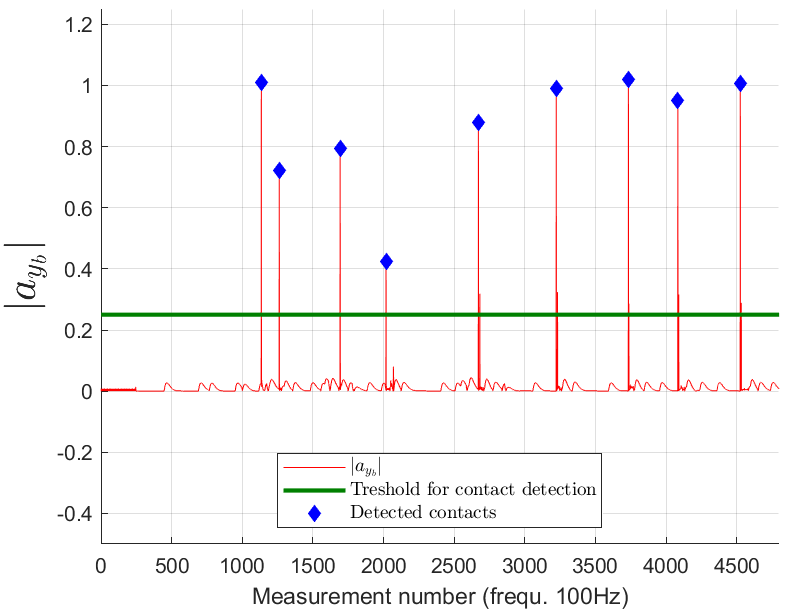,  width = 1\columnwidth}}
  \caption{Contact detection from acceleration measurements.}
  \label{fig:contact_detection}
  \vspace{-0.1cm}
\end{figure}

To perform dot-painting, the quadcopter must impact the wall at each dot coordinates. To do so, for each paint dot $P_i^{*(\rho)}$ to be applied, the chosen strategy is to make use of a PID position controller to first stabilize the quadrotor at a "waiting" position $Q_i^{*(\rho)}$ located at a given offset distance $\Delta$ from  $P_i^{*(\rho)}$ (see Figure \ref{fig:notations}):
\begin{equation*}
	Q_i^{*(\rho)} = P_i^{*(\rho)} - \left[ 0 \quad \Delta \quad 0 \right]^T 
\end{equation*}
 Following this, a velocity controller is then used to move the quadcopter along the $y_\rho$ direction towards the wall, until contact.\\
\indent Collisions between the robot and the wall are detected by monitoring the acceleration measurement along the $y_b$ axis provided by the IMU. If the absolute value of this acceleration measurement \(\left| a_{y_b}\right|\) exceeds a predefined threshold (which was chosen to be \(0.25g\)), it is assumed that the robot just touched the facade. An example of acceleration measurements and contact detections is given in Figure \ref{fig:contact_detection}.\\

\indent Once a contact is detected, the position controller is used again to assign the robot back to the \(Q_i^*\) "waiting" point. Using the information of contact, the estimation process of $(\delta^{(w)}, \gamma)$ is run and the updated coordinates of the next paint dot are computed along with the new yaw reference to align the drone with the wall. The whole process is repeated until the last paint dot is applied onto the wall.\\

\subsection{MAV localization}\label{sec:localization}

\indent For applications in GPS-denied or GPS-perturbed environments, such as in close proximity to facades, specific methods have to be used for localization of the drone. Different kinds of Simultaneous Localization And Mapping (SLAM) methods allow to determine the position of a robot from IMU measurements and vision \cite{SLAM1}.\\
\indent In this paper, the AR Drone 2.0 quadcopter has been chosen as test platform. This choice is motivated  by the fact that it is equipped with an IMU and a monocular camera, which are the sensor suite of interest, and by the availability of software drivers and simulation packages for this drone. 
Indeed, a monocular SLAM based on the Parallel Tracking and Mapping (PTAM) algorithm \cite{Tum1}\cite{Tum3}\cite{Tum2} is provided in the \textit{tum\_ardrone} package used for simulation and is therefore used as localization algorithm. It provides an estimate of the pose of the robot in $\mathcal{R}_\rho$ which will be considered in the control algorithm. This estimate is accurate enough for small speed and short term trajectories but may suffer a drift when used over a long period of time. 
\\
\indent If the number of paint dots is large, the mission will indeed have a long duration. To account for the energetic endurance of the robot, it has been chosen to make the quadcopter land after every $\sim$ 8 minutes of flight. It is assumed that landing is realized at the same location as take-off (automatic return to \textcolor{red}{a} home battery charging or changing station for example, equipped with a visual tag so that the drone is able to automatically land on the same point). This "break" in the mission can also be used to refill the painting system, and is finally taken as an opportunity to reset the PTAM algorithm so that the localization of the drone will not be affected by the long-term drift. Of course the estimation process is not reset and the current value obtained for the estimate $(\hat{\delta}^{(w)}, \hat{\gamma})$ is maintained.\\
%
%
%
%
%
%
%
\indent The monocular PTAM algorithm used in this navigation system also provides a 3D point cloud of the environment. An intuitive idea would be to use it to make a 3D reconstruction of the wall, but in our use case, it is not possible. The PTAM method estimates the position of the camera for mapping positions of points of interest in the space by analyzing and processing the image provided by the camera in real time. To do that, the algorithm needs various textured elements to be present in the image. Because of the very close proximity and contacts to the wall, it is not possible to make the camera of the drone face the wall. In addition, looking at low textured walls would not enable us to get enough points of interest for the computer vision algorithm. To overcome this problem, an angular offset of 90deg has been chosen between the optical axis of the camera and the normal to the wall. In other terms, and according to the notations used in the definition of the reference frames (see Figure \ref{fig:notations}), the camera is directed along the $x_b$ axis of the robot, and the painting system along $y_b$ to be facing the wall. Therefore the camera is not looking directly at the wall. By doing this, the PTAM method is able to detect enough points for localization but the 3D mapping capability of the algorithm may not be used anymore to get 3D information on the wall. This motivates the proposed strategy of using collisions to generate information about the wall without bothering the PTAM method in its process of tracking and mapping.
%
%

\section{Dot painting on a flat wall}

\subsection{Scenario description}

\begin{figure}[!h]
  \vspace{-0.2cm}
  \centering
   {\epsfig{file = 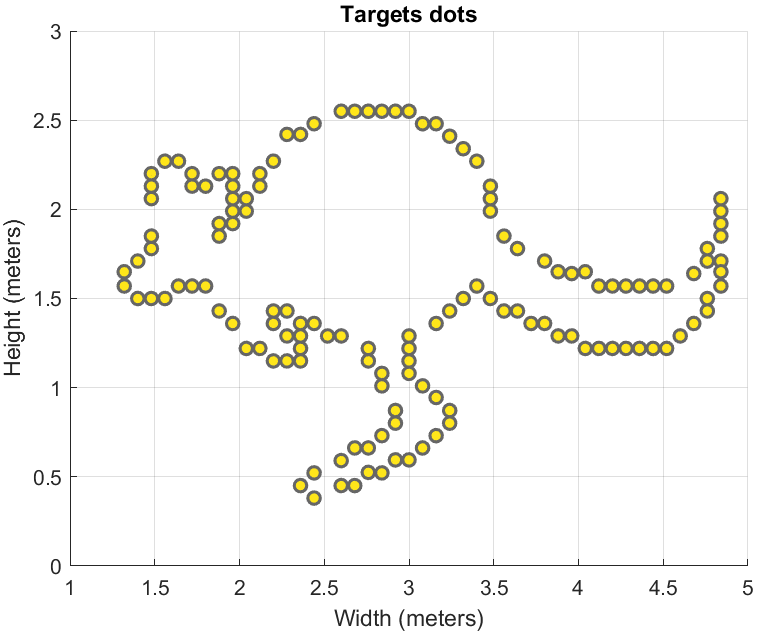, width = 1\columnwidth}}
  \caption{Reference positions of the paint dots to be applied used for the different simulations.}
  \label{fig:targets}
  \vspace{-0.1cm}
\end{figure}

\begin{figure}[!h]
  \vspace{-0.2cm}
  \centering
   {\epsfig{file = 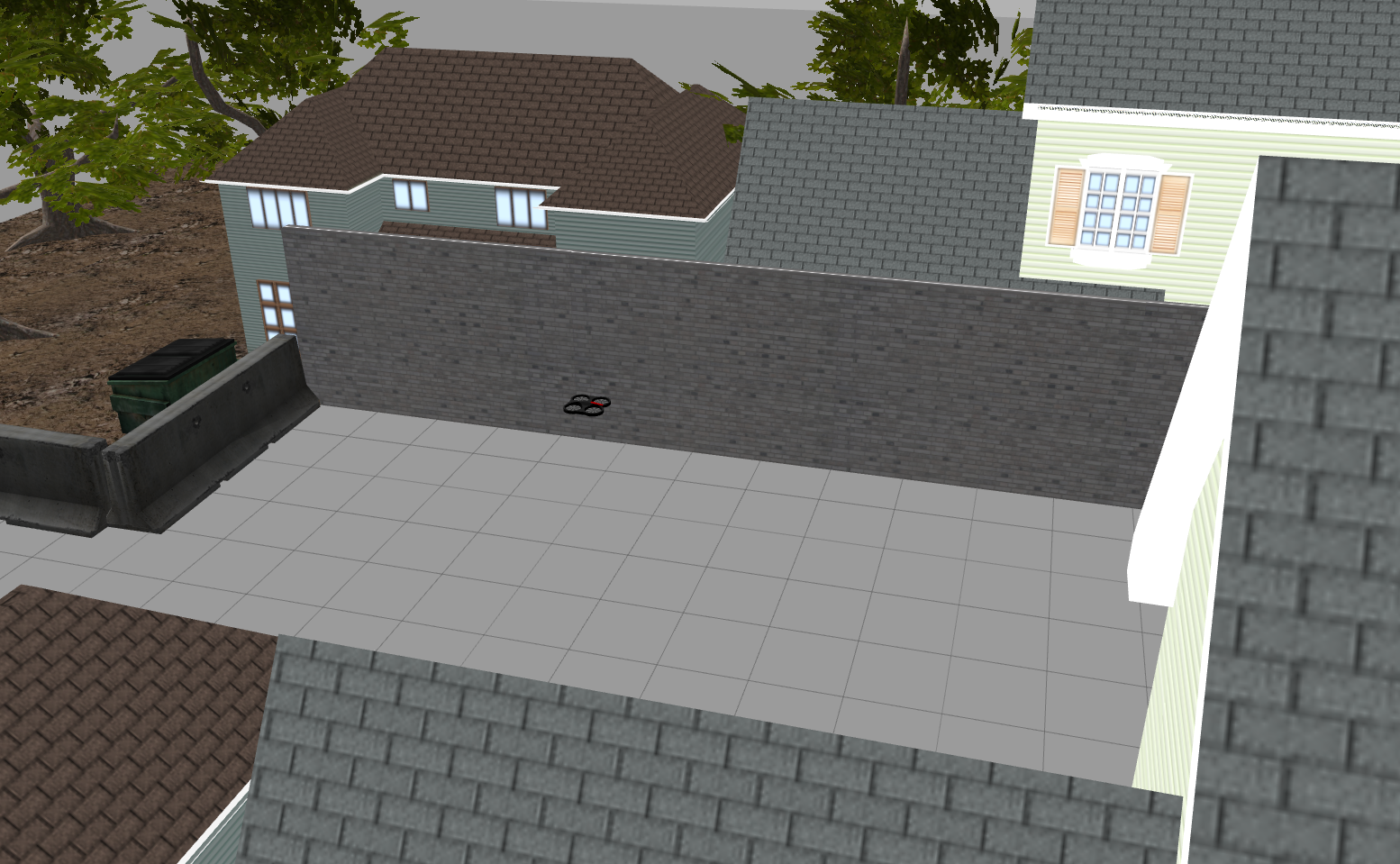, width = 1\columnwidth}}
  \caption{The Gazebo world used for the flat wall scenario.}
  \label{fig:Gazebo1}
  \vspace{-0.1cm}
\end{figure}

\noindent In this scenario, the goal is to apply a set of 140 dots (\( P_{0:140}^{ref}\)) on the flat wall visible in Figure \ref{fig:Gazebo1}, forming a predefined drawing that is 3.5 meters wide and 2.1 meters high (see Figure \ref{fig:targets}). The application order of the dots has been arbitrarily chosen to be from the right to the left and from the top to the bottom. To improve the estimation process of $(\delta^{(w)}, \gamma)$, an additional point is introduced just before the first dot of the drawing and far enough from it. Therefore, contact information from these two first distant points will lead to a rather good first estimate, that will be improved then after collecting more and more contact information. This process is described in more details in the next subsection.
\\

\subsection{Estimation from contact information}
\noindent The objective here is to compute an estimate $(\hat{\delta}^{(w)}, \hat{\gamma})$ of $(\delta^{(w)}, \gamma)$ by making use of the information on the discrepancy between  the reference coordinates \(P_i^{ref(w)}\) of the points to be painted and the obtained coordinates \( P_i^{c(\rho)}\) of the collision points. 
\\
Let us denote by $i$ the index of the last collision point achieved. The estimation process is run after each collision, for $i\geq2$. 
For $i< 2$, the initial value $( \hat{\delta}^{(w)}(0), \hat{\delta}(0) ) $ is used instead. 
\\
The estimate $(\hat{\delta}^{(w)}_i, \hat{\gamma}_i)$ is computed, after the $i$-th contact, as the solution of the optimization problem 
\begin{equation}\label{eq:min_pb}
    \min_{(\bar{\delta}^{(w)}, \bar{\gamma})}{ \sum_{j=1}^{i}{ \omega_j \left\Vert R(\bar{\gamma}) P_j^{c(\rho)} + \bar{\delta}^{(w)} - P_j^{ref(w)} \right\Vert^2  }  }
\end{equation}
%
%
%
where the coefficients $\omega_j$ are some strictly positive weights. Note that a 2D problem is considered here by defining the reference points and contact points used in \eqref{eq:min_pb} by two-dimensional vectors from the $x$ and $y$ components of the 3D points. Other choices, eg. values of $\omega_j$ depending on the localization accuracy at the time of contact $j$, will be investigated in future work.

\indent This is the classical problem of finding the best-fitting rigid transformation between two sets of matching points and for which an analytical solution exists. 
To compute this solution, a method based on Singular Value Decomposition (SVD) is used. The different steps of this method are the one presented in \cite{Sorkine-Hornung17}:\\ \\
1. Compute the centroids of both point sets:\\
\begin{equation}
\bar{P}^{c(\rho)}=\frac{\sum_{j=1}^i\omega_j P_j^{c(\rho)}}{\sum_{j=1}^i\omega_j}
\end{equation}
\begin{equation}
\bar{P}^{ref(w)}=\frac{\sum_{j=1}^i\omega_j P_j^{ref(w)}}{\sum_{j=1}^i\omega_j} \end{equation}
\\
2. Compute the centered vectors ($j=1,..,i$):\\
\begin{equation}
x_j= P_j^{c(\rho)}-\bar{P}^{c(\rho)}
\end{equation}
\begin{equation}
y_j=P_j^{ref(w)}-\bar{P}^{ref(w)}
\end{equation}
3. Compute the \(2 \times 2\) covariance matrix:\\
\begin{equation}
S=XWY^T
\end{equation}\\
where X and Y are the \(2 \times i\) matrices that have \(x_i\) and \(y_i\) as their columns, respectively, and \(W=diag(\omega_1,\omega_2,\dots,\omega_i)\). In our case, the weights \(\omega_i\) have all been set to 1.\\
\\
4. Compute the Singular Value Decomposition \(S=U\Sigma V^T\). The rotation matrix solution of \eqref{eq:min_pb} is given by:\\
\begin{equation}
R(\hat{\gamma}_i)=V\begin{pmatrix} 
1\\
& 1\\
& & .\\
& & & .\\
& & & & 1\\
& & & & & det(VU^T)
\end{pmatrix}U^T
\end{equation}\\
\\
The angle estimate $\hat{\gamma}_i$ is directly determined from the knowledge of $R(\hat{\gamma}_i)$.
\\
\\
5. The translation vector solution of \eqref{eq:min_pb} is given by:\\
\begin{equation}
    \hat{\delta}^{(w)}_i = \bar{P}^{ref(w)}-R(\hat{\gamma}_i)\bar{P}^{c(\rho)}
\end{equation}

\indent From the obtained estimate $( \hat{\delta}^{(w)}_i, \hat{\gamma}_i ) $, the coordinates of the next points \(P_k^{*(\rho)}\), $k>i$, can be updated, as well as the yaw reference $\psi^r$ to be applied to the MAV so that its painting system faces the wall: 
\begin{equation}
    P_k^{*(\rho)}=R(\hat{\gamma}_i) (P_k^{ref(w)}-\hat{\delta_i}^{(w)})\\ \\
\end{equation}
\begin{equation}
\psi^r=-\hat{\gamma}_i \hspace{2.5mm}
\end{equation}
%
%
%
%

\subsection{Results} \label{results1}


Simulation results are provided in this section on the flat wall scenario. The approach proposed in the previous subsection has been applied to different set ups corresponding to several values of the angular offset $\gamma$. Results are first presented for the case $\gamma=$10deg, and performance is then analyzed for $\gamma \in \left\{-15,-10,0,10,15 \right\}$ deg. In all cases, the position offset is chosen as $\delta^{(w)} = \left[0 \, \, -1.1 \, \, 0\right]^T$m. 
\\
To assess of the performance of the proposed approach, a comparison is provided between the following cases: 
\begin{itemize}
    \item online estimation of the offsets $(\delta^{w}, \gamma)$, and localization of the drone based on the PTAM algorithm ("\textit{offset estimation and PTAM}")
    \item online estimation of the offsets $(\delta^{w}, \gamma)$, and localization of the drone based on ground truth ("\textit{offset estimation and GT}")
    \item no estimation of the offsets $(\delta^{w}, \gamma)$, and localization of the drone based on ground truth ("\textit{no offset estimation and GT}")
\end{itemize}

\begin{figure}[!h]
  \centering
   {\epsfig{file = 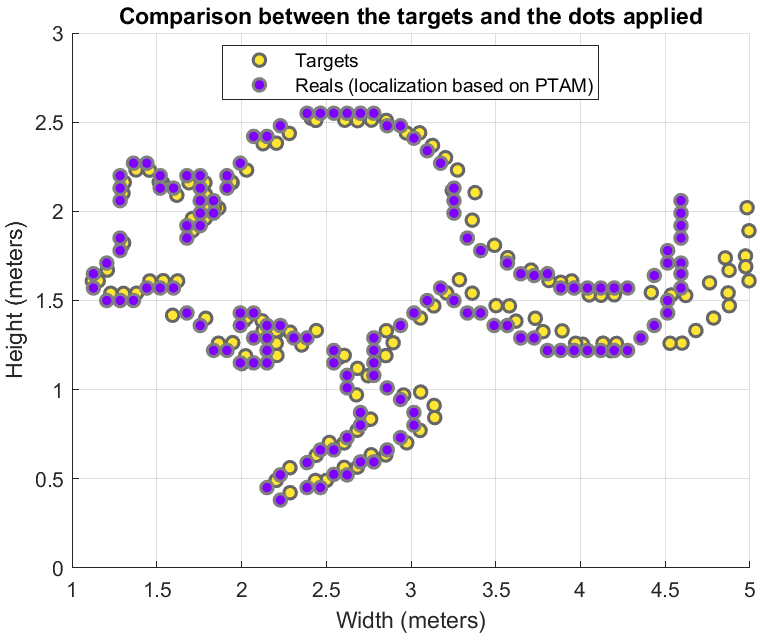, width = 1\columnwidth}}
  \caption{Desired and obtained drawings for the case "\textit{offset estimation and PTAM}" ($\gamma$=10deg)}
  \label{fig:drawing_PTAM_estim}
  \vspace{-0.1cm}
\end{figure}

\begin{figure}[!h]
  \centering
   {\epsfig{file = 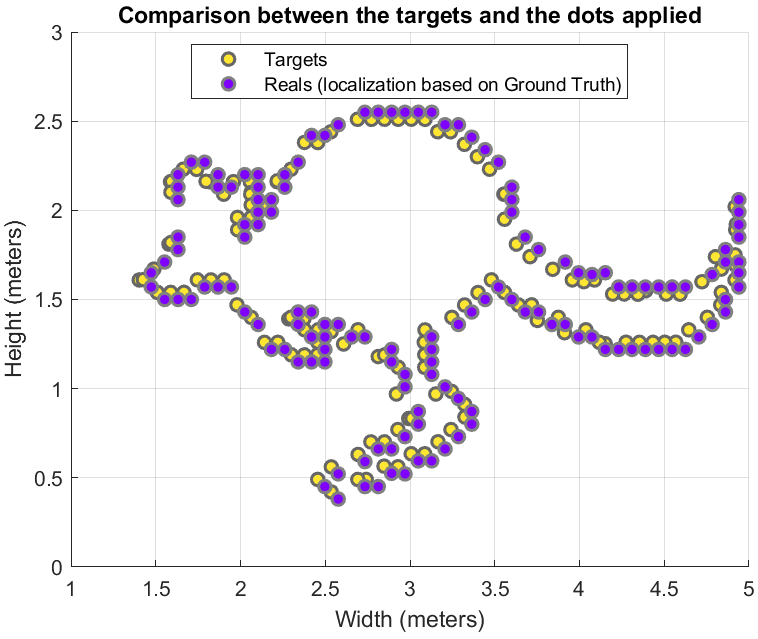, width = 1\columnwidth}}
  \caption{Desired and obtained drawings for the case "\it{offset estimation and GT}" ($\gamma$=10deg)}
  \label{fig:drawing_GT_estim}
  \vspace{-0.1cm}
\end{figure}

\begin{figure}[H]
  \centering
   {\epsfig{file = 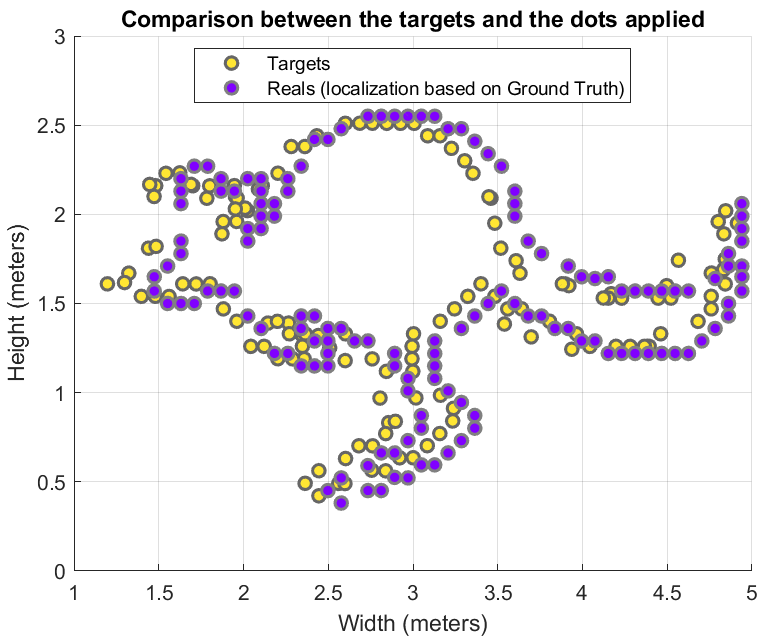, width = 1\columnwidth}}
  \caption{Desired and obtained drawings for the case "\it{no offset estimation and GT}" ($\gamma$=10deg)}
  \label{fig:drawing_GT_no_estim}
  \vspace{-0.1cm}
\end{figure}

\begin{figure}[t]
  \centering
   {\epsfig{file = 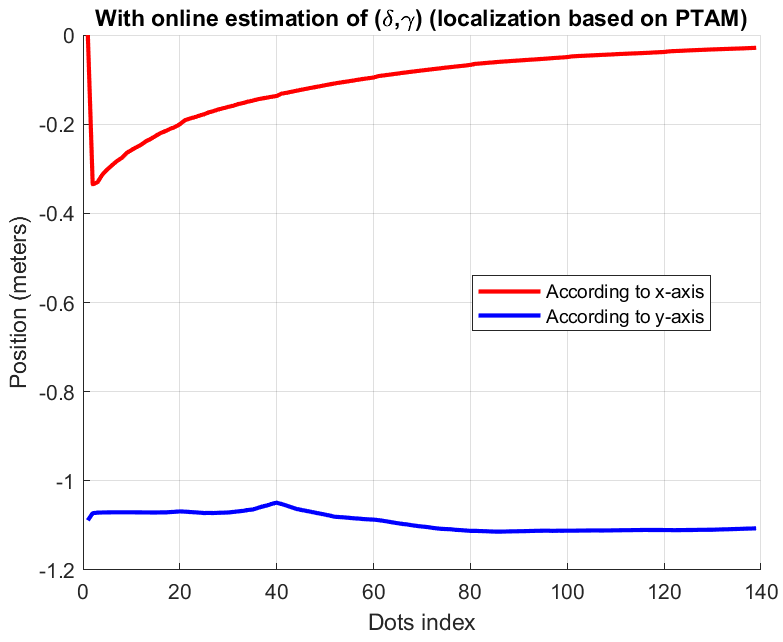, width = 1\columnwidth}}
  \caption{Estimates of \(\delta_x\) and \(\delta_y\) for the case "\textit{offset estimation and PTAM}" ($\gamma$=10deg)}
  \label{fig:delta1}
  \vspace{-0.1cm}
\end{figure}

\begin{figure}[H]
  \centering
   {\epsfig{file = 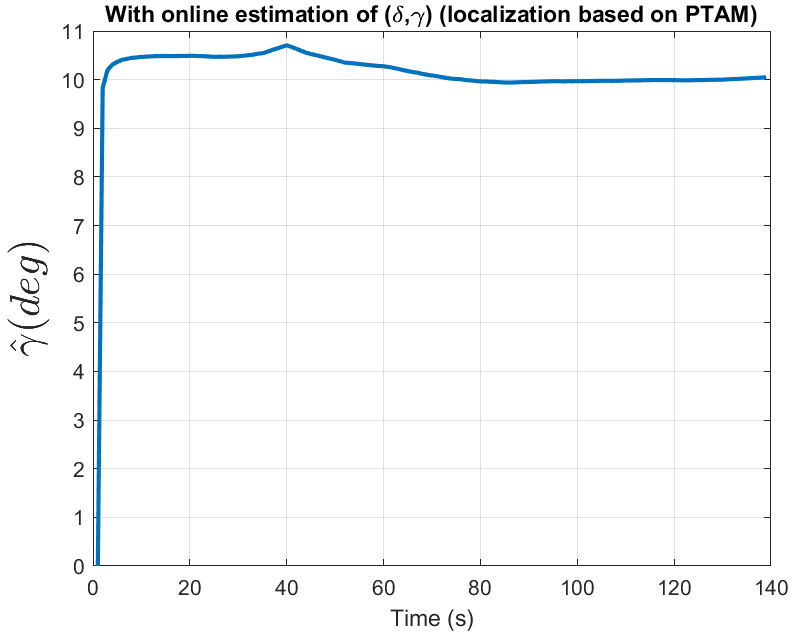, width = 1\columnwidth}}
  \caption{Estimate of \(\gamma\) for the case "\textit{offset estimation and PTAM}" ($\gamma$=10deg)}
  \label{fig:gamma1}
  \vspace{-0.1cm}
\end{figure}

\begin{figure}[b]
  \centering
   {\epsfig{file = 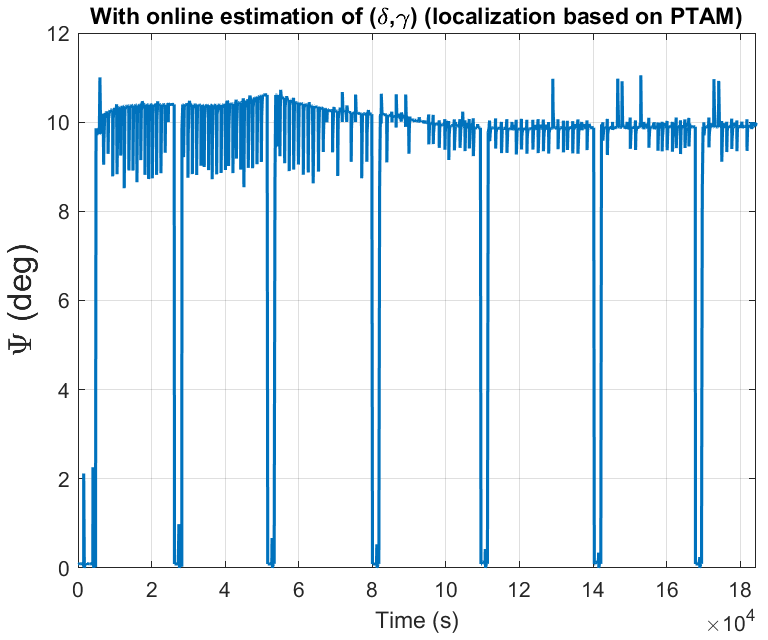, width = 1\columnwidth}}
  \caption{Quadcopter yaw angle for the case "\textit{offset estimation and PTAM}" ($\gamma$=10deg)}
  \label{fig:yaw1}
  \vspace{-0.1cm}
\end{figure}

\begin{figure}[t]
  \centering
   {\epsfig{file = 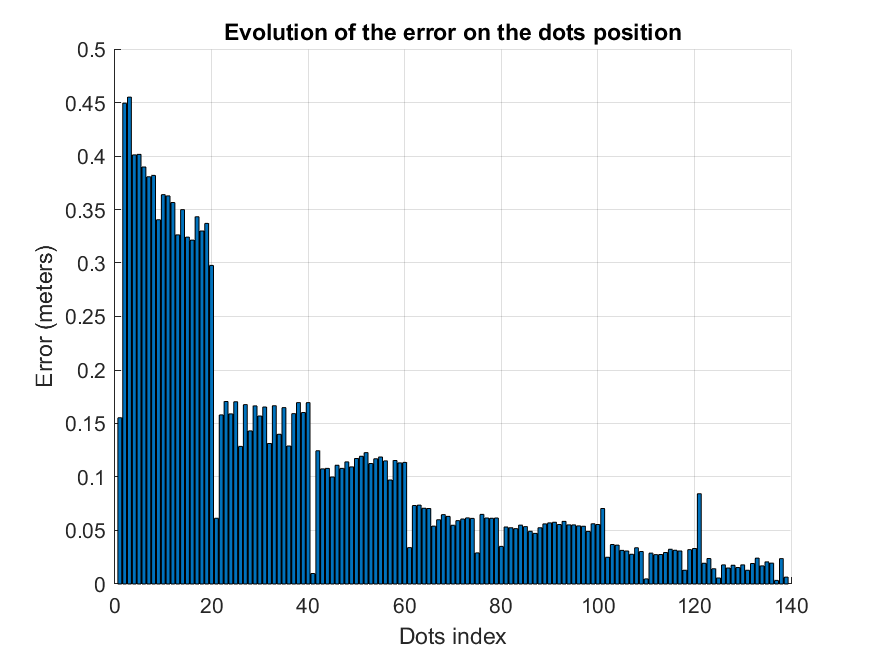, width = 1\columnwidth}}
  \caption{Evolution of the distance error between desired and obtained dot positions for the case  "\textit{offset estimation and PTAM}" ($\gamma$=10deg)}
  \label{fig:RMSE1}
  \vspace{-0.1cm}
\end{figure}

\indent Comparison between the first two cases enable to identify the influence of the localization algorithm. Comparison to the third case enable to analyze the influence of the proposed offset estimation method. In the third case the robot is therefore sent directed to the wall toward the $y_\rho$ direction, whatever the value of $\gamma$. The first case, "\textit{offset estimation and PTAM}", is the one corresponding to "real" conditions and is the sole for which the multiple take-off and landing procedure presented in Section \ref{sec:localization} is realized.
\\
\indent Figure \ref{fig:drawing_PTAM_estim} to \ref{fig:RMSE1} correspond to the particular case $\gamma=10$ deg. %
Figures \ref{fig:drawing_PTAM_estim}, \ref{fig:drawing_GT_estim}, \ref{fig:drawing_GT_no_estim} present a comparison between the desired and the obtained drawings respectively for the three aforementioned cases. Comparison of these figures clearly demonstrates the improvement of the drawing accuracy obtained when applying the proposed estimation method. Influence of the localization precision is also clearly visible, when comparing Figures \ref{fig:drawing_PTAM_estim} and \ref{fig:drawing_GT_estim}, showing a degradation when using the on-board localization algorithm compared to a theoretical localization based on ground truth. Nevertheless, the "\textit{offset estimation and PTAM}" case which corresponds to conditions close to reality results in a good drawing. \\
\\

\indent Figures \ref{fig:delta1} and \ref{fig:gamma1} present the time evolution of the estimates \(\hat{\delta}^{(w)}\) and \(\hat{\gamma}\) respectively, for the case "\textit{offset estimation and PTAM}". As can be seen, the computed estimates converge close to the true values. The evolution of the yaw angle \(\psi\) of the robot is therefore well compensated to make the painting system face the wall. This can be seen on Figure \ref{fig:yaw1} where the yaw angle of the robot is controlled so as to compensate for the angular deviation \(\gamma\). Chattering on the curve is due to contacts with the wall and the points \(\psi=0\) that are visible every each \(\sim\)8 minutes correspond to successive take-offs and landings as explained in Section \ref{sec:localization}. As can be seen on Figure \ref{fig:RMSE1}, the online estimation hence enables to decrease the distance error between desired and obtained dots as the estimation improves over time.\\


Accuracy of the obtained drawings is finally assessed and compared in terms of Root Mean Square Error of the dot positions over the whole mission. These results are provided in Table \ref{tab:opti1} for the three cases and the different tested values of $\gamma$. As can be seen the theoretical and ideal case ("\textit{offset estimation and GT}") shows the best accuracy and the improvement obtained by the proposed approach is clearly visible when compared to the case without online estimation ("\textit{no offset estimation and GT}"). The practical case which relies on PTAM localization introduces more errors, but within an acceptable order of magnitude. As was shown on Figure \ref{fig:drawing_PTAM_estim}, the largest errors occur at the first applied points, due to the time of convergence of this estimation process, and are responsible of these RMSE values.
\\
\begin{table}[!h]
\caption{RMSE (in centimeters) on dot positions.}\label{tab:opti1} \centering
\begin{tabular}{|c||c|c|c|}
  \hline
  & \multicolumn{2}{|c|}{With estimation} & Without  \\
  \hline
  \(\gamma\hspace{1.5mm}\)(degrees) & PTAM & GT & GT\\
  \hline
  0 & 8,09 & 5,67 & 4,81\\
  \hline
  +10 & 15,71 & 4,67 & 15,79\\
  \hline
  -10 & 15,51 & 7,71 & 14,49\\
  \hline
  +15 & 22,68 & 7,71 & 18,89\\
  \hline
  -15 & 23,37 & 11,49 & 17,34\\
  \hline
\end{tabular}

\end{table}

%

\begin{figure}[!h]
  \centering
   {\epsfig{file = 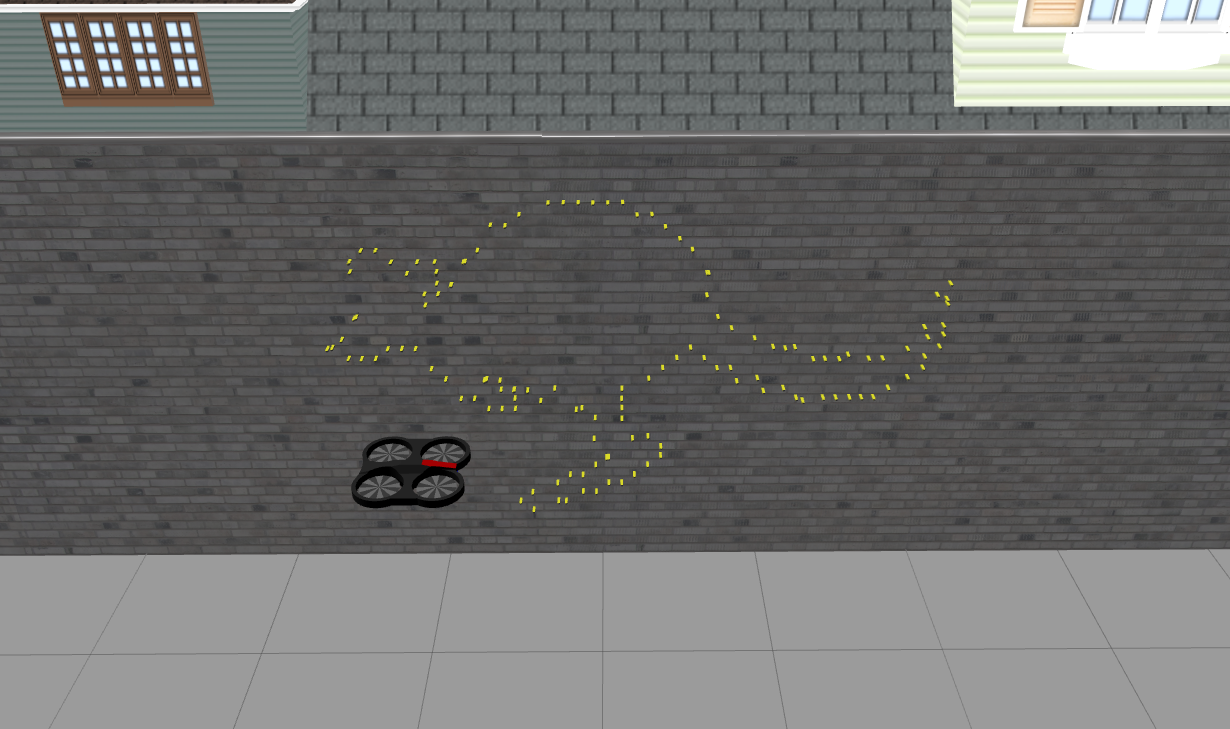, width = 1\columnwidth}}
  \caption{3D illustration of a drawing obtained by the proposed approach}
  \label{fig:drawing_GT_estim_gazebo}
  \vspace{-0.1cm}
\end{figure}

It can be concluded that the proposed approach enables to obtain a good drawing, despite uncertainty on the initial positioning of the robot. A 3D visualization of such drawing is provided for illustration purpose on Figure \ref{fig:drawing_GT_estim_gazebo} for this flat wall scenario. 

\section{Dot painting on multiple segments wall}

\subsection{Scenario description}
\begin{figure}[!h]
  \vspace{-0.2cm}
  \centering
   {\epsfig{file = 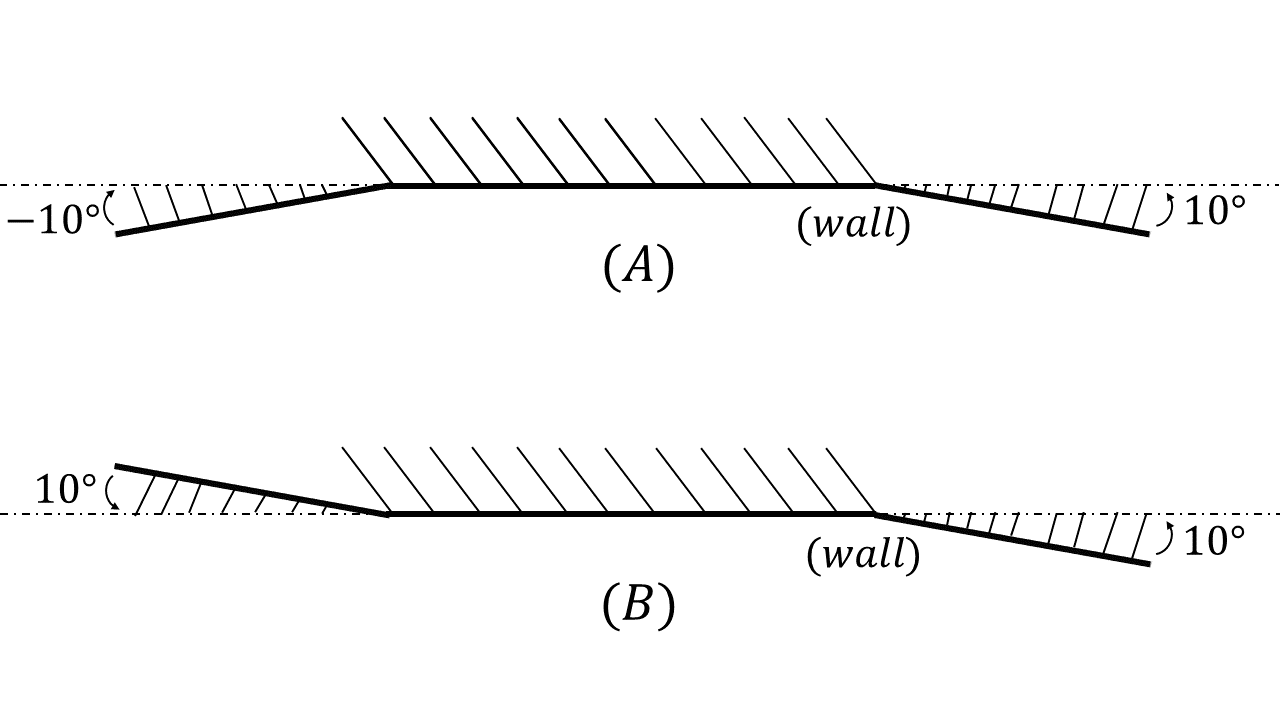, width = 1\columnwidth}}
  \caption{Example of a 3 segments wall with angular offsets $\gamma_i$ $\left\{-10,0,+10\right\}$deg (A) and $\left\{+10,0,+10\right\}$deg (B).}
  \label{fig:3segments}
  \vspace{-0.1cm}
\end{figure}

\begin{figure}[H]
  \vspace{-0.2cm}
  \centering
   {\epsfig{file = 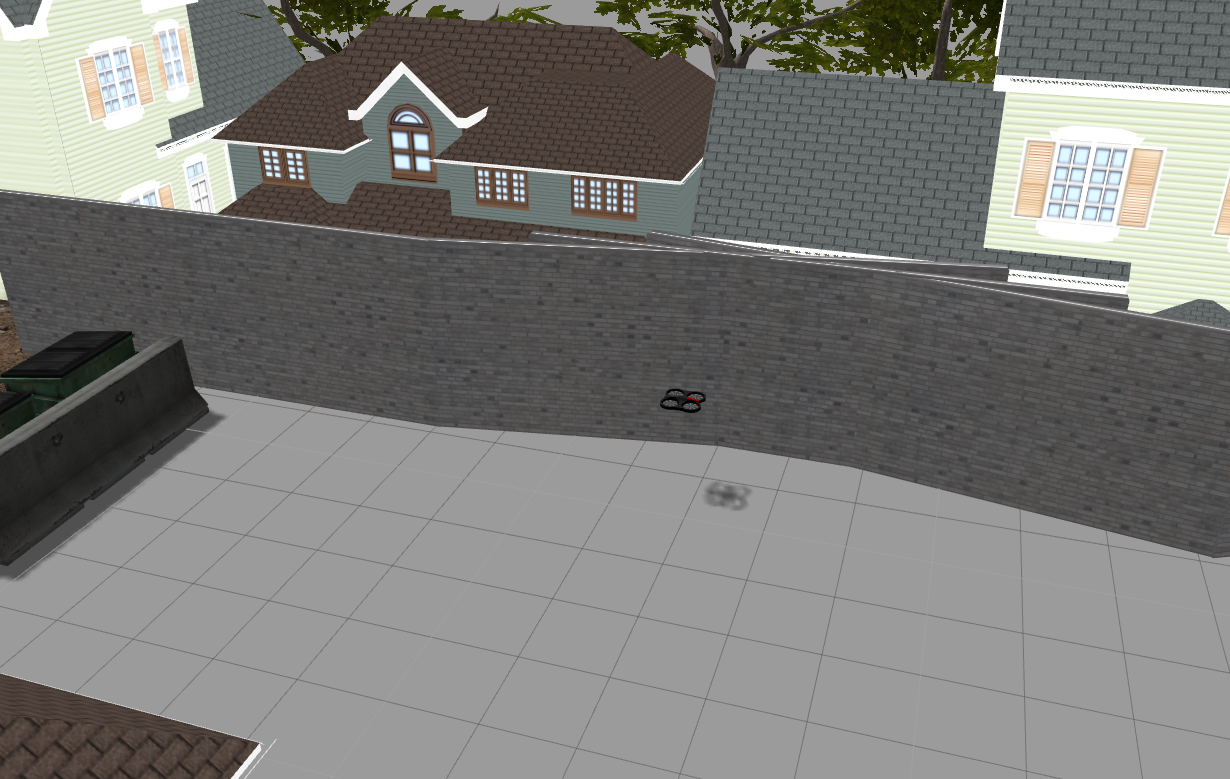, width = 1\columnwidth}}
  \caption{The Gazebo world used for this scenario.}
  \label{fig:Gazebo2}
  \vspace{-0.1cm}
\end{figure}

\noindent In this scenario, the goal is to apply the same set of paint dots \( P_i^{ref}\) as in the previous flat wall scenario but on a multiple segments wall this time. Each of them is associated with an offset \(\gamma_i\) which characterizes the orientation difference between the facade and the theoretical position of the quadcopter (see Figure \ref{fig:3segments}). An extension of the previous approach is proposed, and tested on a scenario with a wall consisting of three segments (Figure \ref{fig:Gazebo2}).

\subsection{Proposed approach}

\noindent This approach consists of estimating the shape of the surface by online computing, as previously, the angle offset between the surface and the quadcopter after each dot application. It is then possible to determine the optimal command to make the robot orthogonal to the surface model for the next dots.\\
\indent From the above point of view, we consider the shape \(\Sigma\) of a multiple segments wall as a set of lines\,\(L_i\):
\begin{equation}
\Sigma = \{ L_1,L_2,\dots,L_n\},\text{with}\hspace{1mm}L_i: y^{(\rho)}=a_ix^{(\rho)}+b_i
\end{equation}
\indent After applying the first two dots, we start by computing the first line \(L_1\) that goes through them. This gives an initial estimate of the shape of the wall. For each new dot $i$ applied, the discrepancy is computed between the real coordinates of where the contact point actually occurred (\(P_i^{c(\rho)}\)) and the predicted one corresponding to (\(P_i^{ref(\rho)}\)) by using the previously estimated shape of the wall: 
\begin{equation}
    \epsilon_i=\left\| a_{i-1}x_i^{ref(\rho)}+b_{i-1}-y_i^{c(\rho)}\right\|
\end{equation}\\
where $L_{i-1}: y^{(\rho)}=a_{i-1}x^{(\rho)} + b_{i-1}$ corresponds to the the shape model of the current part of the facade on which the robot is working, as estimated from previous contacts.  If the discrepancy $\epsilon_i$ exceeds a predetermined threshold \(\Bar{\epsilon}\), it is considered that the current dot was applied on a different segment of the facade with an offset in the \(\gamma\) angle compared to the previous segment. 
In this case, a new line $L_i: y^{(\rho)}=a_ix^{(\rho)}+b_i$ is computed by using the coordinates of the contact point $i$ and the ones of the previous contact point $i-1$. This line is added to \(\Sigma\). If $\epsilon_i$ is lower than the threshold \(\Bar{\epsilon}\), the new line $L_i$ added to $\Sigma$ is computed by performing a linear regression including all previous contact points fitting this model. 
\\

\indent As for the flat wall case, once this estimation process has been done, the coordinates of the next contact point is updated by taking into account the estimate of the current wall segment. This time, the update is done only for the $y_{i+1}^{*(\rho)}$ coordinate of the next point $P_{i+1}^{*(\rho)}$
\begin{equation}
    y_{i+1}^{*(\rho)} := a_i x_{i+1}^{*(\rho)} + b_i
\end{equation}
This choice is made arbitrarily to ensure no "\textit{longitudinal}" deformation of the drawing (i.e. no deformation along the width), meaning that someone standing in front of the symmetry axis of the wall (if any) and starring at the drawing would not notice any deformation. On the contrary, the observed deformation will increase with the distance to this specific observation position. This is different from a fresco application where all parts of the drawing should be projected normally to each corresponding wall segment. 

\subsection{Results}

Simulation results are provided in this section on the multiple segments wall scenario. The approach proposed in the previous subsection has been applied to different setups corresponding to several values of wall angular offsets $\gamma$. Results are first presented for the case $\gamma=-10_{deg} | 0_{deg} | +10_{deg}$.

\begin{figure}[!h]
  \vspace{-0.2cm}
  \centering
   {\epsfig{file = 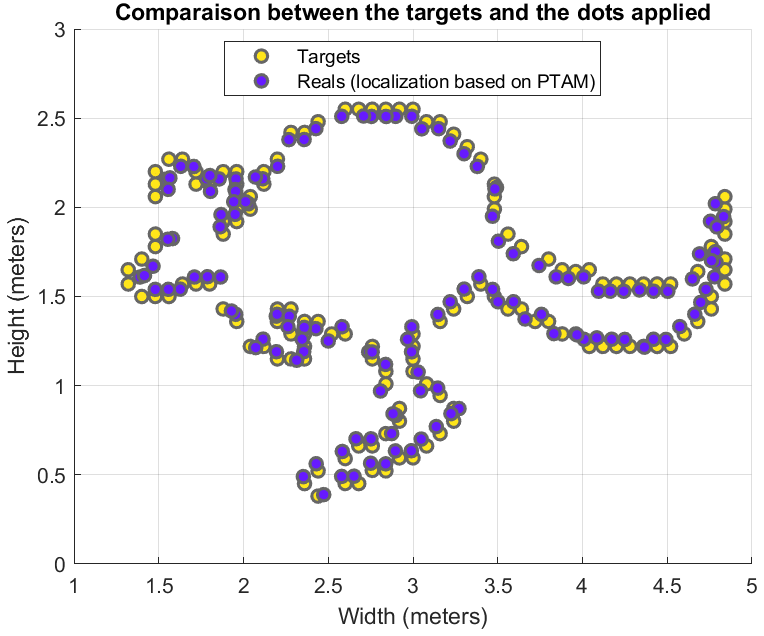, width = 1\columnwidth}}
  \caption{Desired and obtained drawings for the case "\textit{offset estimation and PTAM}" ($\gamma=-10_{deg} | 0_{deg} | +10_{deg}$).}
  \label{fig:trace1}
  \vspace{-0.1cm}
\end{figure}

\begin{figure}[!h]
  \centering
   {\epsfig{file = 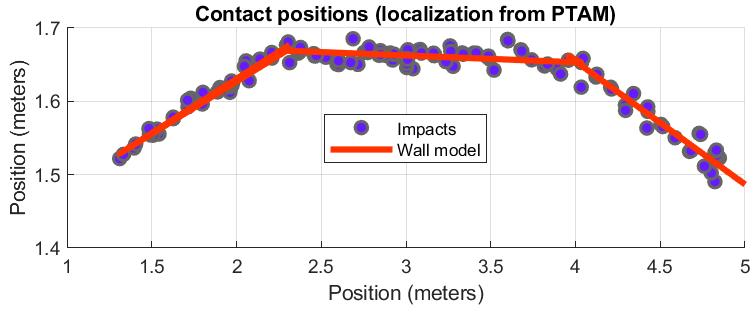, width = 1\columnwidth}}
  \caption{Model of the wall made from contact information and using the localization based on PTAM.}
  \label{fig:wallmodel}
  \vspace{-0.1cm}
\end{figure}

\begin{figure}[!h]
  \vspace{-0.2cm}
  \centering
   {\epsfig{file = 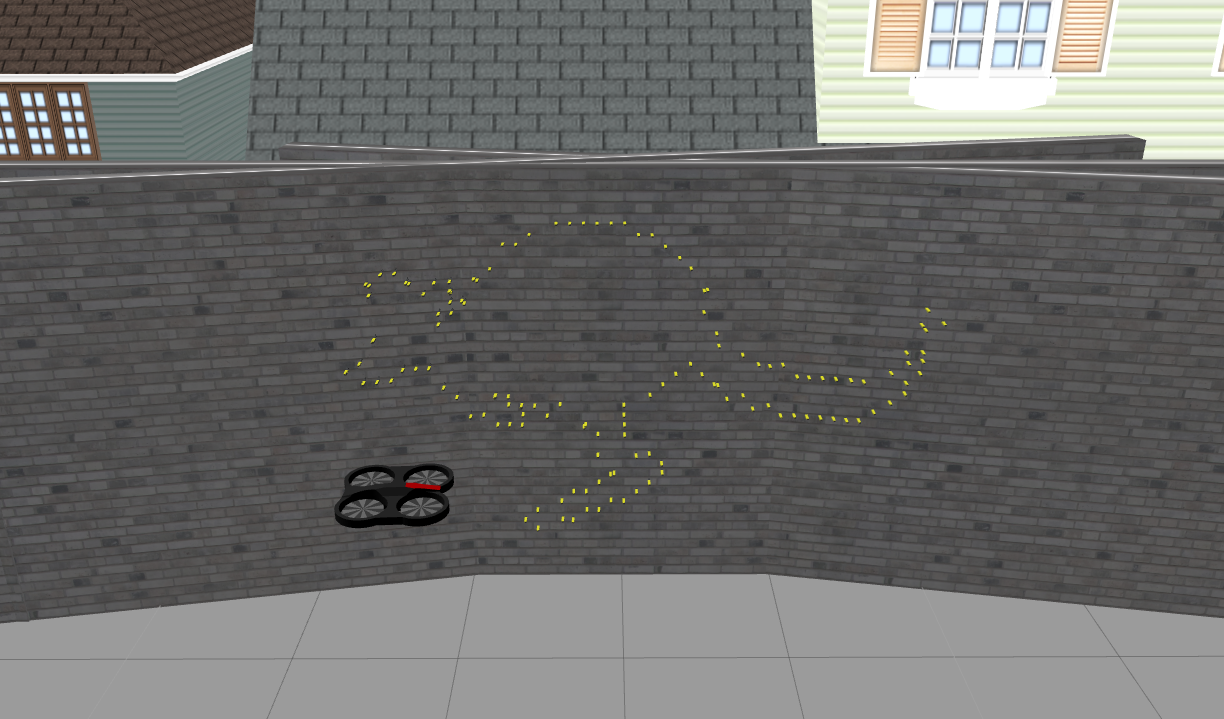, width = 1\columnwidth}}
  \caption{3D illustration of a drawing obtained by the proposed approach.}
  \label{fig:results2}
  \vspace{-0.1cm}
\end{figure}

Accuracy of the obtained drawings is assessed and compared again in terms of Root Mean Square Error of the dot positions over the whole mission. These results are provided in Table \ref{tab:opti2} for the three cases and the different values of $\gamma$. As shown on Figure \ref{fig:wallmodel}, the wall model generated is very similar to the real one and a good drawing is obtained (see Figure \ref{fig:results2}).\\

\begin{table}[!h]
\caption{RMSE (in centimeters) on dot positions.}\label{tab:opti2} \centering
\begin{tabular}{|c||c|c|c|}
  \hline
  & \multicolumn{2}{|c|}{With estimation} & Without  \\
  \hline
  \(\gamma\hspace{1.5mm}\)(degrees) & PTAM & GT & GT\\
  \hline
  -10 \(|\) 0 \(|\) +10 & 3,74 & 4,96 & 4,68\\ 
  \hline
  -15 \(|\) 0 \(|\) +15 & 6,75 & 5,91 & 4,59\\ 
  \hline
  +10 \(\hspace{0.7mm}|\) 0 \(|\) +10 & 7,70 & 5,85 & 5,03\\ 
  \hline
  -10 \(|\) 0 \(|\) -10 & 5,01 & 5,72 & 4,63\\ 
  \hline
\end{tabular}
\end{table}

\section{\uppercase{Conclusion}}
In this paper, an estimation and control method has been proposed for the problem of dot painting by a quadrotor Micro Air Vehicle. Based only on on-board sensors, this method enables us to deal with uncertainties on the initial positioning of the drone with respect to the wall and with uncertainties on the shape of the wall. Making use of information of contacts between the MAV and the wall, the proposed online estimation procedure compensates for positioning errors due to such uncertainties and results in an improvement of the positioning accuracy of the paint dots. Performance analysis has been proposed in terms of accuracy of the drawings obtained for different simulation scenarios.
\\ 
Future work will focus on flight experiments of the proposed approach.

\newpage
\bibliographystyle{apalike}
{\small
\bibliography{icinco2019_biblio}}

\begin{thebibliography}{}

\bibitem[Briod et~al., 2013]{Briod}
Briod, A., Kornatowski, P., Klaptocz, A., Garnier, A., Pagnamenta, M.,
  Zufferey, J.-C., and Floreano, D. (2013).
\newblock Contact-based navigation for an autonomous flying robot.
\newblock In {\em IEEE/RSJ Int. Conf. on Intelligent Robots and Systems}.

\bibitem[Engel et~al., 2012a]{Tum3}
Engel, J., Sturm, J., and Cremers, D. (2012a).
\newblock Accurate figure flying with a quadrocopter using onboard visual and
  inertial sensing.
\newblock In {\em Workshop on Visual Control of Mobile Robots, at the IEEE/RSJ
  Int. Conf. on Intelligent Robot Systems}.

\bibitem[Engel et~al., 2012b]{Tum2}
Engel, J., Sturm, J., and Cremers, D. (2012b).
\newblock Camera-based navigation of a low-cost quadrocopter.
\newblock In {\em IEEE/RSJ Int. Conf. on Intelligent Robots and Systems}.

\bibitem[Engel et~al., 2014a]{Tum1}
Engel, J., Sturm, J., and Cremers, D. (2014a).
\newblock Scale-aware navigation of a low-cost quadcopter with a monocular
  camera.
\newblock {\em Robotics and Autonomous Systems}, 62(11):1646--1656.

\bibitem[Engel et~al., 2014b]{Tum}
Engel, J., Sturm, J., and Cremers, D. (2014b).
\newblock {\em TUM ARDrone}.
\newblock \url{http://wiki.ros.org/tum_ardrone}.

\bibitem[Galea et~al., 2016]{Dot2}
Galea, B., Kia, E., Aird, N., and Kry, P.~G. (2016).
\newblock Stippling with aerial robots.
\newblock Symposium on Computational Aesthetics / Expressive, 2016.

\bibitem[Galea and Kry, 2017]{Dot1}
Galea, B. and Kry, P.~G. (2017).
\newblock Tethered flight control of a small quadrotor robot for stippling.
\newblock In {\em IEEE/RSJ Int. Conf. on Intelligent Robots and Systems}.

\bibitem[Handy-Paint-Products, 2014]{Handy}
Handy-Paint-Products (2014).
\newblock {\em The World's First Drone Painting Company}.
\newblock
  \url{https://diydrones.com/profiles/blogs/the-world-s-first-drone-painting-company}.

\bibitem[Leigh et~al., 2016]{Pantograph}
Leigh, S., Agrawal, H., and Maes, P. (2016).
\newblock A flying pantograph: Interleaving expressivity of human and machine.
\newblock In {\em 10th Conf. on Tangible Embedded and Embodied Interaction}.

\bibitem[Mei and Rives, 2007]{SLAM1}
Mei, C. and Rives, P. (2007).
\newblock Cartographie et localisation simultan\'ee avec un capteur de vision.
\newblock In {\em Journ\'ees Nationales de la Recherche en Robotique}.

\bibitem[Monajjemi, 2014]{AR}
Monajjemi, M. (2014).
\newblock {\em AR Drone Autonomy}.
\newblock \url{http://wiki.ros.org/tum_ardrone}.

\bibitem[Most, 2017]{Misha}
Most, M. (2017).
\newblock {\em Evolution2.1}.
\newblock \url{http://www.mishamost.com/exhibitions-1/2017/10/10/evolution21}.

\bibitem[Ollero et~al., 2018]{Franchi2}
Ollero, A., Heredia, G., Franchi, A., Antonelli, G., Kondak, K., Sanfeliu, A.,
  Viguria, A., de~Dios, J.~M., Pierri, F., Cortés, J., Santamaria-Navarro, A.,
  Trujillo, M., Balachandran, R., Andralde-Cetto, J., and Rodriguez, A. (2018).
\newblock The aeroarms project: Aerial robots with advanced manipulation
  capabilities for inspection and maintenance.
\newblock {\em IEEE Robotics and Automation Magazine}, 25(4).

\bibitem[Pagliarini and Hautop~Lund, 2009]{Pagliarini2009}
Pagliarini, L. and Hautop~Lund, H. (2009).
\newblock The development of robot art.
\newblock {\em Artificial Life and Robotics}, 13(2):401--405.

\bibitem[Sorkine-Hornung and Rabinovich, 2017]{Sorkine-Hornung17}
Sorkine-Hornung, O. and Rabinovich, M. (2017).
\newblock Least-squares rigid motion using svd.
\newblock Technical report, Department of Computer Science, ETH Zurich.

\bibitem[Yüksel et~al., 2019]{Franchi1}
Yüksel, B., Secchi, C., Bülthoff, H.~H., and Franchi, A. (2019).
\newblock Aerial physical interaction via ida-pbc.
\newblock {\em International Journal of Robotics Research}.

\end{thebibliography}

\end{document}